\documentclass[conference]{IEEEtran}

\usepackage{times}
\usepackage[numbers]{natbib}
\usepackage[bookmarks=true]{hyperref}

	\usepackage[latin1]{inputenc}   
	\usepackage[T1]{fontenc}        
	\usepackage[margin=0.8in]{geometry}
	\usepackage{scrextend}
	\usepackage[para]{bigfoot} \DeclareNewFootnote[para]{default}
	\usepackage[bottom]{footmisc}
	\usepackage{flushend}
	\usepackage{amsmath,amssymb}
	\usepackage{cancel}

	\usepackage{amsthm}
	\usepackage{siunitx}
	\usepackage{graphicx}
	\usepackage[x11names, rgb]{xcolor}
	\usepackage{tikz}
	\usepackage{pgfplots}
	\usetikzlibrary{automata,arrows,positioning,calc,shapes,fit,shapes.geometric,snakes,shadows.blur}
	\usepackage{algorithm}
	\usepackage[noend]{algpseudocode}
	\algnewcommand{\LineComment}[1]{\vspace{.1em} \item[\(\blacktriangleright\)] #1}
	\let\labelindent\relax
	\usepackage{enumitem}
	\usepackage{booktabs}
	\usepackage{siunitx}
	\usepackage{enumitem}

\begin{document}

\title{Model-free Grasping with Multi-Suction Cup Grippers for Robotic Bin Picking}

\author{\authorblockN{Philipp Schillinger\authorrefmark{1},
		Miroslav Gabriel\authorrefmark{1},
		Alexander Kuss\authorrefmark{1},
		Hanna Ziesche\authorrefmark{1}, and
		Ngo Anh Vien\authorrefmark{1}
	}
	\\
	\authorblockA{\authorrefmark{1}Bosch Center for Artificial Intelligence, Renningen, Germany.\\
		Email: firstname.lastname@de.bosch.com}}

\maketitle

\begin{abstract}
	This paper presents a novel method for model-free prediction of grasp poses for suction grippers with multiple suction cups.
	Our approach is agnostic to the design of the gripper and does not require gripper-specific training data.
    In particular, we propose a two-step approach, where first, a neural network predicts pixel-wise grasp quality for an input image to indicate areas that are generally graspable. 
    Second, an optimization step determines the optimal gripper selection and corresponding grasp poses based on configured gripper layouts and activation schemes.
    In addition, we introduce a method for automated  labeling for supervised training of the grasp quality network.
    Experimental evaluations on a real-world industrial application with bin picking scenes of varying difficulty demonstrate the effectiveness of our method.
\end{abstract}

\IEEEpeerreviewmaketitle

\section{Introduction}
\begin{figure}[t]
	\includegraphics[width=.99\linewidth,trim=0 30 0 20,clip]{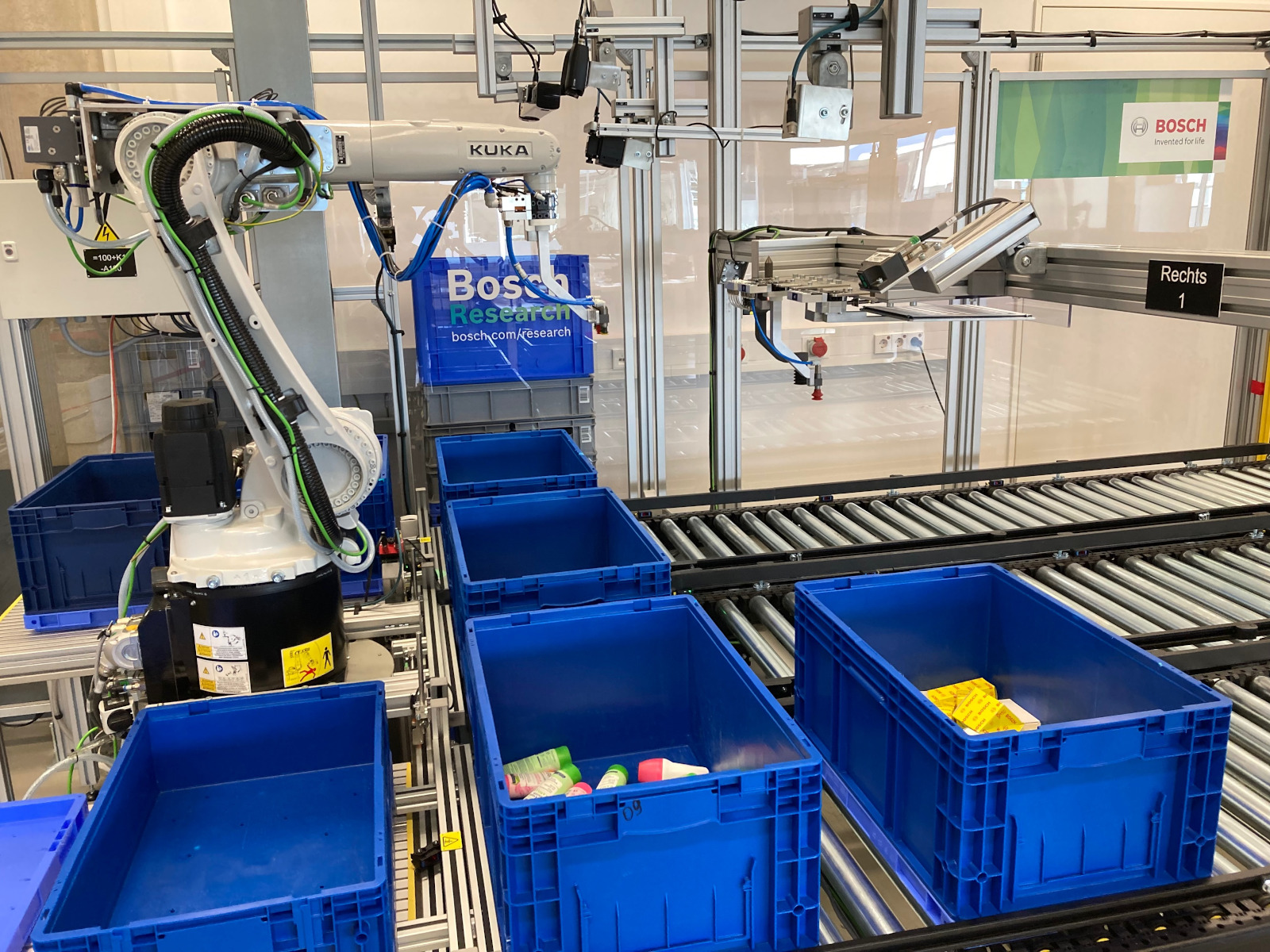}
	\caption{Robotic bin picking system with six-axis industrial robot, overhead RGB-D camera, tool changer, single-suction gripper, multi-suction gripper, bins containing diverse objects and conveyor belt.}
	\label{fig:bin_picking_setup}
	\vspace*{-1em}
\end{figure}

Model-free grasping with multi-suction grippers is a key challenge for fully automating many pick-and-place tasks in industry and logistics.
Fig.~\ref{fig:bin_picking_setup} shows a typical robotic bin picking system for warehouse order fullfilment including an industrial robot equipped with a multi-suction gripper, an overhead RGB-D camera and bins containing diverse objects positioned on a conveyor belt.
Recent research proposes machine learning methods that enable model-free grasp prediction for a wide variety of unseen objects in unstructured environments \cite{newbury2022deep, kleeberger2020survey}.

Most approaches focus on grasp prediction for parallel-jaw grippers or single-suction grippers.
However, suction grippers with multiple suction cups are rarely studied so far, although they are capable of balancing torques which is favorable for dynamic movements of objects with large dimensions \cite{mantriota2007optimal}. 
They also enable lifting of heavier objects without damaging their surfaces by distributing the required grasp force over multiple suction cups.
Some multi-suction grippers allow the activation of different suction cup groups to offer flexibility in diverse object portfolios.
However, multi-suction grippers are more challenging for grasp prediction approaches to deal with due to their complex geometry and alternative activation patterns.

In this paper, we propose a method for model-free prediction of grasp qualities for suction grippers that is independent of the actual gripper design, e.g., the number and size of suction cups.
More specifically, we present a gripper-agnostic grasp quality prediction including a procedure for automatic labeling and supervised training, allowing for generalization to new object shapes.
Furthermore, we present a method for optimal gripper selection and rotation based on the inferred, gripper-agnostic pixel-wise grasp quality prediction combined with footprint images to represent specific gripper design configurations.

Our contributions are as follows:
(1) Method for model-free prediction of grasp qualities agnostic to the suction gripper design, including automated labeling for supervised training.
(2) Method for optimal gripper selection and orientation by matching of arbitrary suction gripper footprints.

We evaluate contribution (1) by a comparison of our proposed pixel-wise grasp quality prediction with existing methods based on various bin picking scenes of different levels of difficulty.
In addition, we demonstrate contribution (2) by real-world experiments of performing multi-suction cup grasp prediction on an industrial bin picking application.

\begin{figure*}[t]
	\includegraphics[width=.99\linewidth]{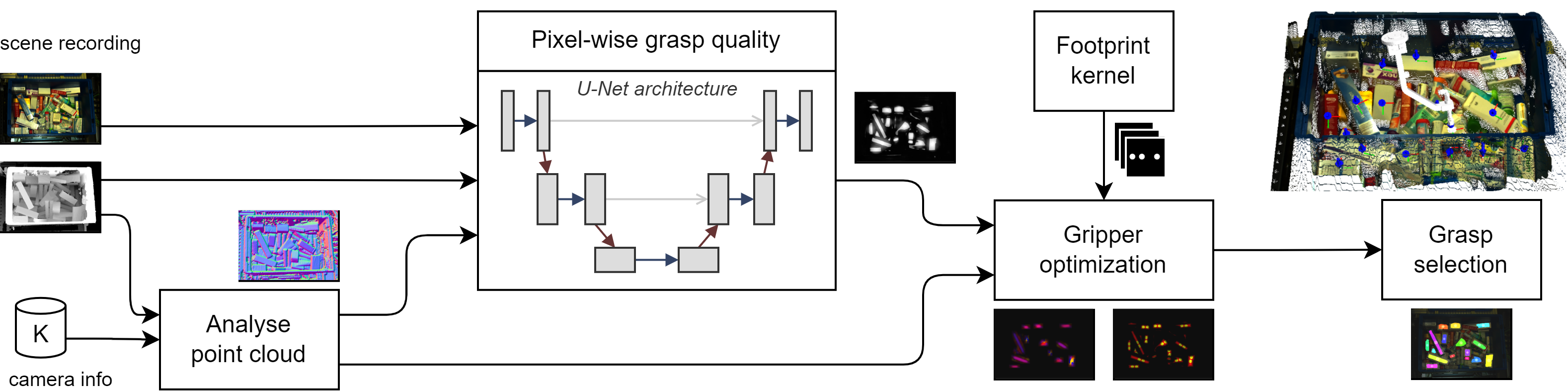}
	\caption{Summary of our proposed approach. Our method receives an RGB-D scene recording and camera information to derive a pixel-wise grasp quality. Afterwards, by providing gripper geometry information as footprints, a gripper optimization finds the best poses and corresponding grippers.}
	\label{fig:approach}
\end{figure*}

\section{Related Work}
Modern robot grasping methods often use deep learning techniques trained on large datasets to predict grasps. 
Most grasp prediction approaches rely on some form of predicting a pixel-wise grasp quality map that represents the grasp success probability at each pixel.
\citet{mahler2017dex} and
\citet{zeng2022robotic} propose to learn a grasp map for suction and parallel-jaw grasps from a supervised dataset using RGB-D or depth as input.
Following this approach, GG-CNN \cite{MorrisonLC18} and FC-GQ-CNN \cite{satish2019policy} propose to predict a grasp quality map and a 4~DoF parallel-jaw grasp configuration for each pixel.
\citet{cao2021suctionnet} propose a pixel-wise grasp map and and a grasp configuration prediction or single-suction grippers.
\citet{BreyerCOSN20} propose to generate voxel-wise grasps using a truncated signed distance function for a parallel-jaw gripper.
A follow-up work optimizes grasp prediction jointly with object shape reconstruction \cite{0002ZSFZ21}. 
Other approaches use point clouds as input and predict point-wise grasp qualities and gripper configurations \cite{ZhaoZLWT021, qin2020s4g, yang2021robotic, jeng2020gdn, ni2020pointnet++,fang2020graspnet, li2020learning}.

Various gripper designs are available for robotic grasping. Parallel-jaw and suction grippers are commonly used and effective for regular object sets \cite{correll2016analysis,hernandez2016team}. In bin picking scenarios, suction grippers have an advantage over parallel-jaw grippers for object reachability. Dex-Net 3.0 \cite{mahler2017suction} improves success rates of Dex-Net 2.0 \cite{mahler2017dex} by training a grasp quality neural network specifically for suction grippers. \citet{shao2019suction} propose a self-supervised learning method for simulated suction-based picking. Suctionnet-1billion predicts grasps for single-suction grippers through end-to-end training of a pixel-wise prediction network \cite{cao2021suctionnet}. \citet{jiang2022learning} jointly learn pixel-wise grasp quality and robot reachability maps for suction vacuum cups. \citet{zeng2018robotic}, winners of the Amazon picking challenge, propose learning a pixel-wise grasp map for a hybrid gripper combining parallel-jaw and suction cup functions.

Although there are advanced multi-suction gripper designs in industrial products \cite{maggi2022introducing, mantriota2007optimal}, there is little research on explicitly modelling and using them for robotic grasping.
Recent efforts focus on optimizing a single network for the prediction of grasps for different gripper types \cite{khargonkar2022neuralgrasps,shao2020unigrasp,xu2021adagrasp}.
However, no prior work has explored learning a pixel-wise grasp map that can be used for both single- and multi-suction grippers without altering the network architecture or the training process.

\section{Problem Statement}

Given an image of the scene and a set of multi-suction grippers, our goal is to predict a set of feasible grasp poses which can be used to transfer arbitrary objects from a source to a target bin.
In particular, our method receives an RGB-D image as input and infers grasp poses from a multi-channel grasp map where each channel per gripper type encodes pixel-wise grasp quality and rotations.

We denote by $S$ an RGB-D scene image of size $\mathbb{R}^{4\times h\times w}$, and by $g = (x, y, z, \alpha, \beta, \gamma, t)$ a grasp configuration predicted at position $(x, y, z)$ with orientation $(\alpha, \beta, \gamma)$ and gripper $t \in T$.
As grippers, we assume that there is a set $T$ of different types of multi-suction grippers that can be used for a grasp.
The problem of predicting multi-suction grasps is to find a mapping $f: S \mapsto g$ for every input image $S$.

We propose a two-step approach for solving this problem as illustrated in Fig.~\ref{fig:approach}.
First, a neural network predicts a pixel-wise "graspability" property for the image, denoting how well each individual pixel can be grasped and in the following referred to as grasp quality.
Second, an optimization step determines the best gripper and corresponding grasp pose based on the predicted grasp quality.

In contrast to methods that directly approximate $f$, this two-step approach has the benefit that no gripper-specific training data is needed.
To the best of our knowledge, a grasp dataset of multi-suction cups does not exist in literature and, in particular, not for our specific multi-suction cup grippers.

\section{Model-free Grasp Quality Prediction}\label{sec:graspability}

The first part of our proposed method consists of a generic, pixel-wise prediction of graspable surfaces.
This prediction can be obtained for a wide range of unknown objects and does not require gripper geometry-specific information.
As input, we use high-resolution RGB-D images, e.g. from industry-grade cameras like \emph{Zivid Two} or \emph{Photoneo PhoXi}, and assume that the intrinsic camera parameters are known.
The output is a pixel-wise grasp quality prediction $Q$ where each pixel ranges from $0$ (not graspable) to $1$ (perfectly graspable) and indicates how suitable the respective pixel is for attaching a suction cup.

\subsection{Grasp Quality Inference}\label{sec:inference}

\begin{figure}
	\centering  
	\includegraphics[width=.42\linewidth,trim=5 0 5 17,clip,angle=90]{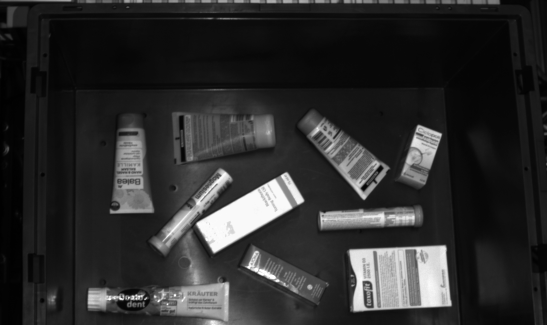}
	\includegraphics[width=.42\linewidth,trim=5 0 5 17,clip,angle=90]{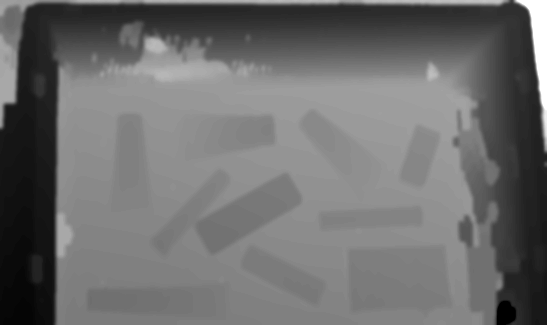}
	\includegraphics[width=.42\linewidth,trim=5 0 5 17,clip,angle=90]{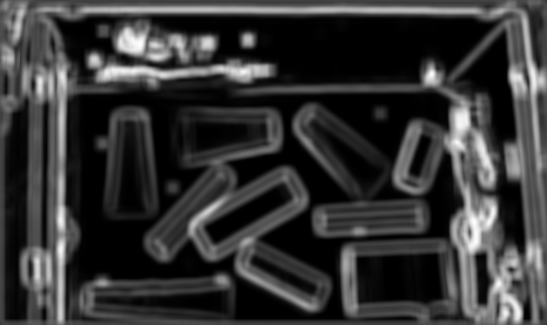}
	\hspace{1em}
	\includegraphics[width=.42\linewidth,trim=5 0 5 17,clip,angle=90]{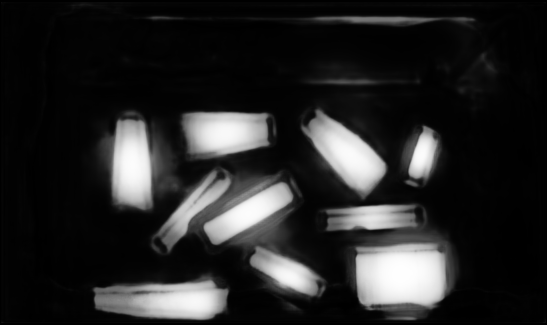}\\
	$($\hspace{1.3em}$S_\mathrm{Gray}$\hspace{1.1em},\hspace{1.3em}$S_\mathrm{Depth}$\hspace{1.1em},\hspace{1.6em}$S_\mathrm{Std}$\hspace{1.5em}$)$\hspace{0.3em}$\mapsto$\hspace{2.3em}$Q$\hspace{2em}
	\caption{Input to the grasp quality network is given by a grayscale channel $S_\mathrm{Gray}$, a depth channel $S_\mathrm{Depth}$, and the standard deviations of surface normals $S_\mathrm{Std}$. Output is a single-channel rating $Q$ how well a suction gripper can be attached to each pixel.}
	\label{fig:channels}
	\vspace*{-0.5em}
\end{figure}

We use a U-Net~\cite{ronneberger2015u} architecture with a ResNet-34~\cite{he2016deep} encoder and a single-channel output $Q$ to infer the pixel-wise grasp quality.
The network input is a three-channel image, $I := (S_\mathrm{Gray}, S_\mathrm{Depth}, S_\mathrm{Std})$, which consists of grayscale $S_\mathrm{Gray}$, depth $S_\mathrm{Depth}$, and the standard deviation of the surface normal vectors $S_\mathrm{Std}$.
Fig.~\ref{fig:channels} shows an example of each channel and the resulting network output $Q$.

Using a single grayscale channel instead of three RGB channels largely retains texture information, but reduces the number of channels that provide this information and, more application-specific, prevents the network from overfitting to colored bins as background.
Using $S_\mathrm{Std}$ as a third channel is motivated by the fact that graspability for suction grippers highly depends on the local surface structure.
If the surface is very irregular, a suction cup is less likely to form a sealed vacuum at that point.

To obtain $S_\mathrm{Std}$, we first calculate the ordered point cloud from $S_\mathrm{Depth}$ and the configured camera matrix $K$ for each pixel $(u, v) \in h \times w$ as
\begin{align*}
	S_\mathrm{Pts}(u, v) = K^{-1} \big(S_\mathrm{Depth}(u, v) [u, v, 1]^\mathrm{T}\big)
\end{align*}
and derive the pixel-wise surface normals, $S_\mathrm{Normals}$. 
Then, $S_\mathrm{Std}$ is computed for a small neighborhood and normalized to a value range $S_\mathrm{Std}(u,v) \in [0,1], \forall u,v$.

One practical challenge in calculating $S_\mathrm{Normals}$ is that this calculation is susceptible to errors and inaccuracies in the depth image.
To address this, we employ two pre-processing steps where the first one fills missing pixels with depth approximations and the second one reduces noise resulting from outlier pixels.

\subsection{Labeling and Supervised Training}\label{sec:training}

For supervised training of the grasp quality network, we require a dataset that contains RGB-D input data $S$, as well as pixel-wise ground-truth for grasp quality $Q^*$.
Annotation of $Q^*$ requires a high effort if done manually for cluttered scenes with many objects or complex geometry.
We therefore choose an alternative approach for obtaining approximate labels, $L$, such that $L(u,v) \approx Q^*(u,v)$.

This automatic labeling approach is based on the insight that for singulated objects with simple geometries, the suitability for a pixel to be grasped inversely correlates with the $S_\mathrm{Std}$, similar to our initial motivation for using $S_\mathrm{Std}$ as an input to the network.
Consequently, we calculate one component of the labels by $L_\mathrm{Std} := 1 - S_\mathrm{Std}$.

Furthermore, we expect grasps closer to the center of mass of an object to be more stable and thus, these pixels should receive a higher grasp quality.
To include this property in the training labels, we cluster pixels of graspable surfaces as given by $L_\mathrm{Std}$ and calculate a second component of the labels $L_\mathrm{Dist}$ as the distance to the respective cluster center.

One limitation of this labeling method is that also the bin and other non-object geometries of the scenes may be considered as graspable areas, solely depending on their surface.
For training, we can circumvent this issue by recording a background image of an empty bin, i.e., a scene recording without any objects before placing the training objects in the scene.
We then use background subtraction based on depth to mask all non-object pixels $M_\mathrm{bg}$ and only consider non-zero labels for object pixels.

\begin{figure}[b]
	\centering  
	\includegraphics[width=.32\linewidth,trim=0 0 0 0,clip]{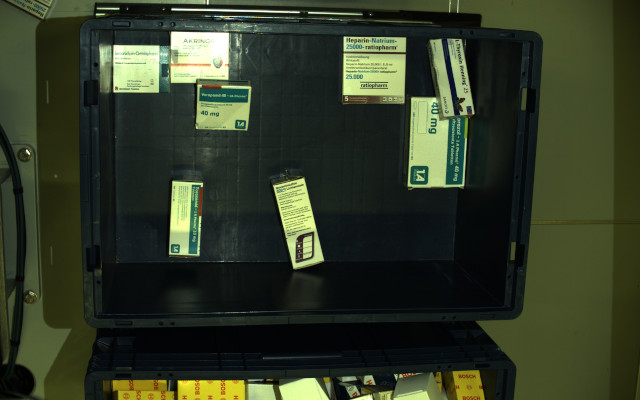}
	\includegraphics[width=.32\linewidth,trim=0 0 0 0,clip]{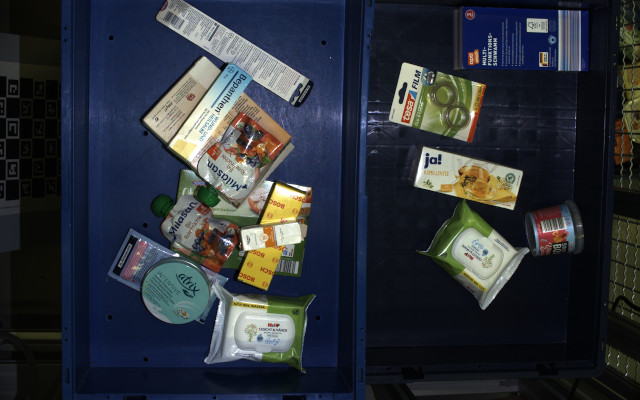}
	\includegraphics[width=.32\linewidth,trim=0 0 0 0,clip]{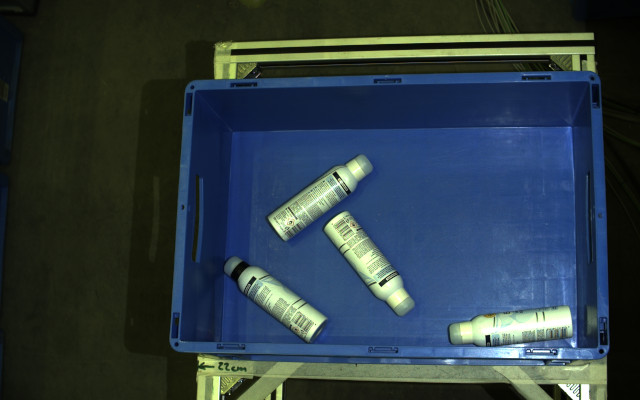}\\[.3em]
	\includegraphics[width=.32\linewidth,trim=0 0 0 0,clip]{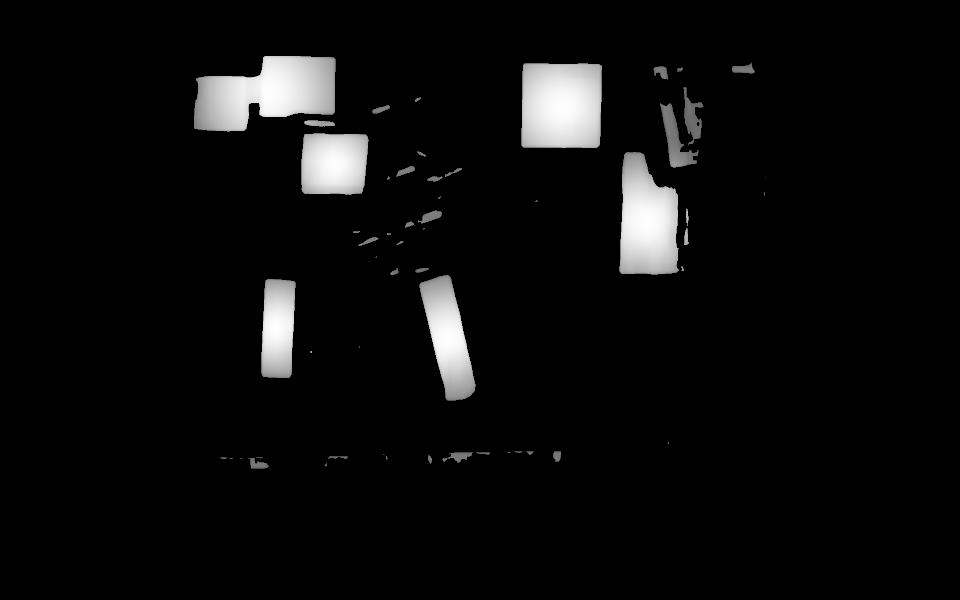}
	\includegraphics[width=.32\linewidth,trim=0 0 0 0,clip]{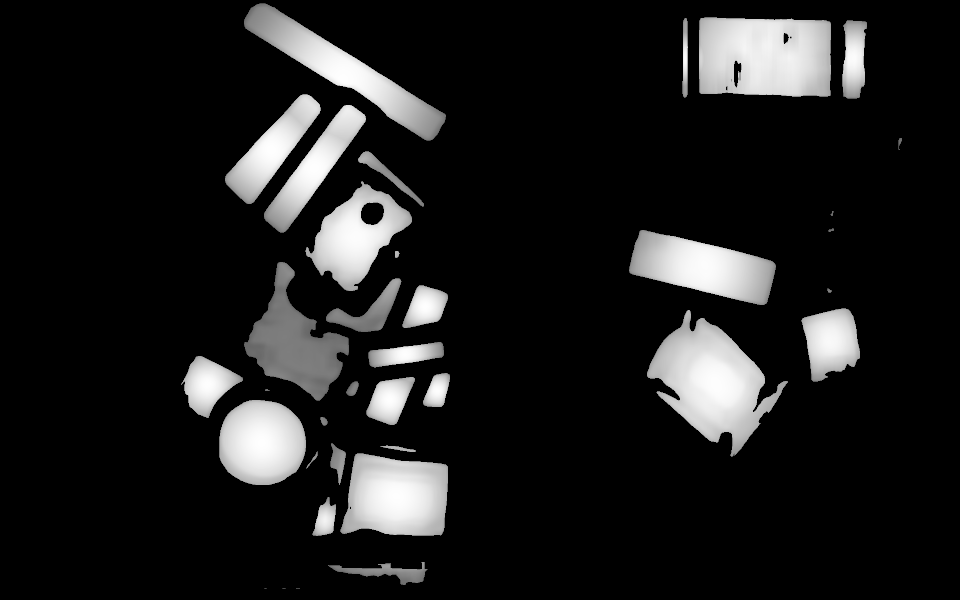}
	\includegraphics[width=.32\linewidth,trim=0 0 0 0,clip]{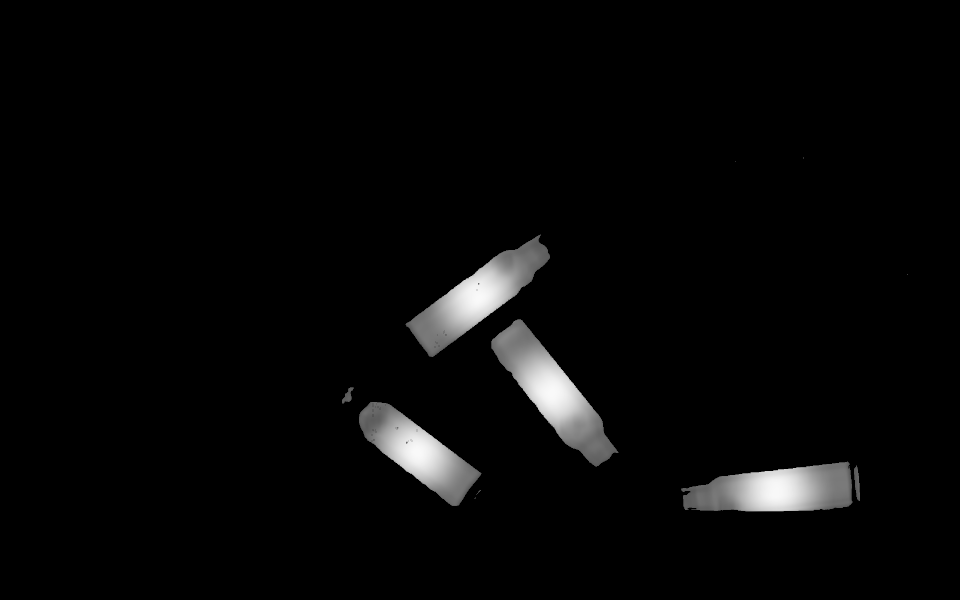}\\
	\caption{Three training examples, RGB input shown top and generated labels $L$ shown in the bottom row. Labels become worse for more complex geometries. Missing labels mainly result from invalid depth information.}
	\label{fig:labels}
\end{figure}

The grasp quality labels $L$ for training are thus given by
\begin{align}
	L = \begin{cases}
	w_\mathrm{Std} L_\mathrm{Std} + w_\mathrm{Dist} {L}_\mathrm{Dist} & \text{where}\, M_\mathrm{bg} > 0\\
	0 & \text{otherwise}
	\end{cases}
	\label{eq:labels}
\end{align}
where the weights $w_\mathrm{Std}$ and ${w}_\mathrm{Dist}$ can balance the influence of the different label components,
but are chosen to be equal in our experiments. The quality metric defined in Eq.~\eqref{eq:labels} distinguishes itself from the approach presented in \cite{jiang2022learning} by using an unnormalized ${L}_\mathrm{Dist}$ score and excluding the residual error of local plane fitting at each pixel, making Eq.~\eqref{eq:labels} more computationally efficient.

Fig.~\ref{fig:labels} shows three examples from our training dataset.
While we observe that this labeling approach works sufficiently well for objects with simple geometries in scenes with only a few instances,
the approximation $L(u,v) \approx Q^*(u,v)$ becomes significantly worse for complex geometries and scenes with object instances that are close together.
Consequently, we limit training data to such simpler scenes that allow for a good approximation.

The approach can be extended to scenes with a larger number of objects or overlapping objects if instance mask annotations are available, which are much easier to obtain by manual annotation than grasp quality labels.
In that case, a background image and object clustering is not required since the background is given by all pixels which are not included in any of the object masks and all other labeling steps can be performed in the same way, including the computation of $S_\mathrm{Std}$ and the final labels according to Eq.~\eqref{eq:labels}.

During training of the grasp quality prediction, the loss $\mathcal{L}$ for some predicted grasp quality $Q$ and target labels $L$ is given by a pixel-wise mean-squared error
\begin{align}
\mathcal{L} = w_\mathrm{bg} \textit{MSE}\left(M_\mathrm{bg} \circ E) + w_\mathrm{fg} \textit{MSE}\big((1 - M_\mathrm{bg}) \circ E\right)
\end{align}
for some prediction error $E := L - Q$ and $\circ$ denoting pixel-wise multiplication.
The background mask $M_{bg}$ balances the loss received for on-object pixels with the one for background pixels with weights $w_\mathrm{bg}$ and $w_\mathrm{fg}$ calculated per image to reflect the ratio of background and foreground.

For the experiments in this paper, we generated a dataset of around 2,000 proprietary recordings of bin picking scenes collected across various robotic cells with a mixed object portfolio similar to those shown in Fig.~\ref{fig:labels}.
Training was then performed for $30$ epochs on a \emph{Nvidia V100} GPU with a batch size of $16$ and images being down-scaled to a resolution of $1280 \times 800$ pixels.
We used stochastic gradient descent with an initial learning rate of $10^{-4}$ and a cosine annealing schedule implemented in \emph{PyTorch}.

\section{Grasp Pose Detection}\label{sec:detection}

The second part of our proposed method is deriving the full grasp pose from the pixel-wise grasp quality prediction described in Sec.~\ref{sec:graspability}.
For this we assume minor application knowledge about the type of available grippers which is manually specified as gripper footprints.
This does not include further scene understanding or context knowledge such as robot kinematics, bin dimensions, or object models.

\subsection{Gripper Footprint Matching}\label{sec:matching}

\begin{figure}[t]
	\centering
	\includegraphics[width=.9\linewidth]{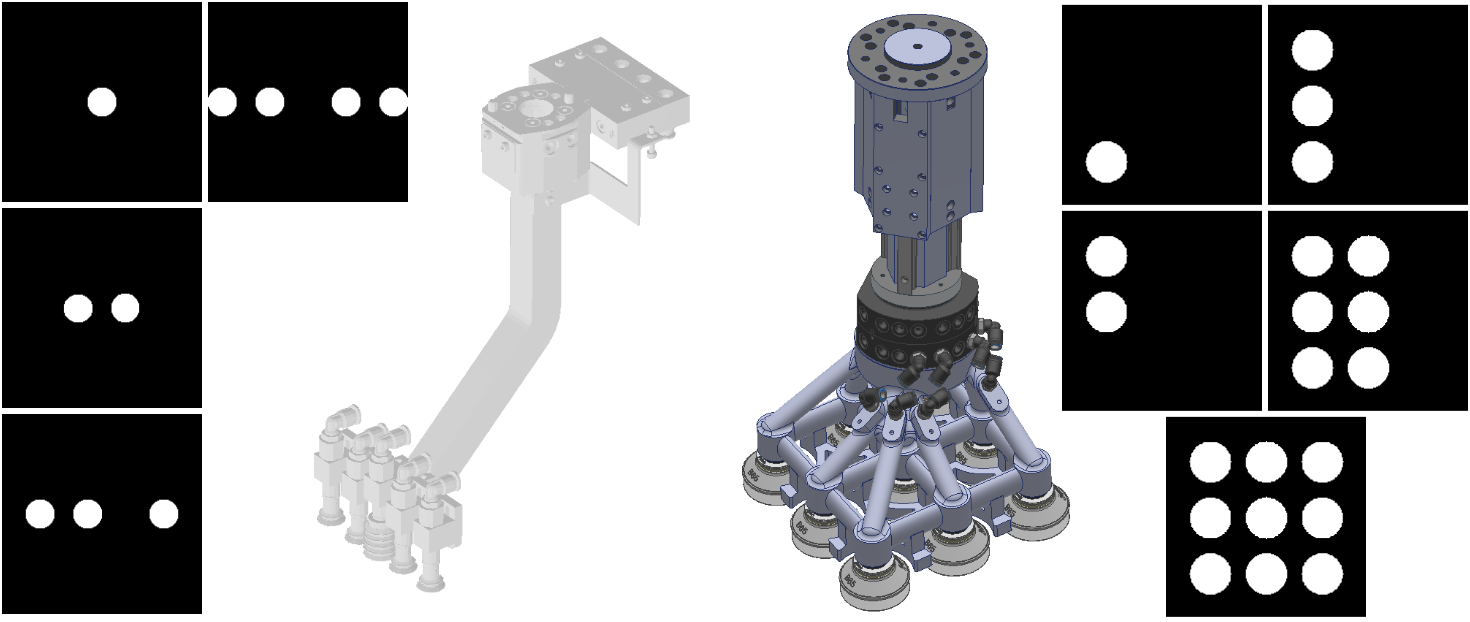}
	\caption{Example footprints of two different multi-suction grippers that both provide multiple activation patterns to choose from. White areas denote full surface contact, black areas mean no contact.}
	\label{fig:footprints}
	\vspace*{-.5em}
\end{figure}

We propose a gripper selection and matching based on specified gripper footprints, such as the ones shown in Fig.~\ref{fig:footprints}.
In this work, we assume that a footprint is always centered at the end-effector pose and that the footprint size is scaled to match the correct pixel-per-mm resolution, for example in our experiment application around two pixels per \SI{}{\milli\meter}.

To identify grasp poses, including a selection of the best gripper and its orientation, we perform a convolution over the inferred grasp quality $Q$.
For this, $n_\mathrm{r}$ different discrete rotation steps of $n_\mathrm{f}$ different gripper footprints are encoded as separate channels in one combined convolution kernel $F \in \mathbb{R}^{n_\mathrm{r} \cdot n_\mathrm{f} \times h_F \times w_F}$ of size $h_F \times w_F$.
The result of a convolution of $Q$ with $F$ is thus a multi-channel pixel-wise prediction how well each gripper type in each rotation can perform a successful grasp at the respective pixel.

\begin{figure}[b]
	\vspace*{-1em}
	\centering
	\includegraphics[width=.49\linewidth,trim=100 35 160 90,clip]{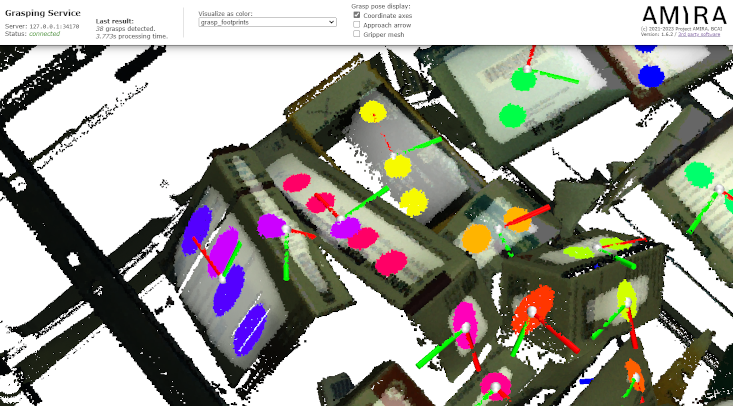} 
	\includegraphics[width=.49\linewidth,trim=115 45 150 85,clip]{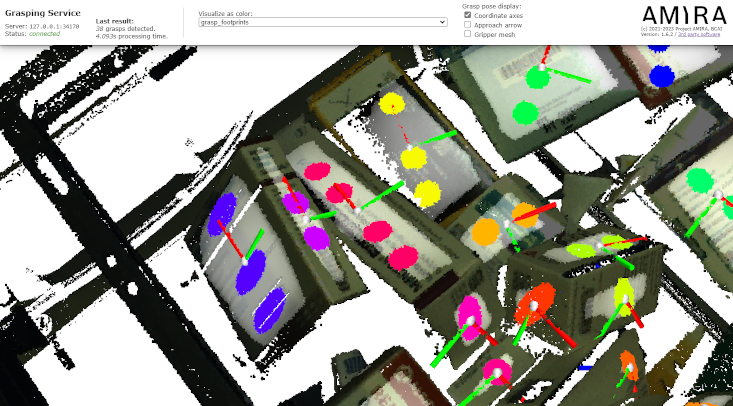}\\
	\caption{Example for the edge wrapping issue. For the purple grasp in the left image, a large three-suction cup gripper is incorrectly selected and ignores scene geometry. Instead, a smaller footprint aligned with the object surface would be correct and is the result when applying the proposed method, as shown in the right image.}
	\label{fig:footprint_edge_wrapping}
\end{figure}

\begin{figure*}[t]
\includegraphics[width=.19\linewidth,trim=75 0 100 50,clip]{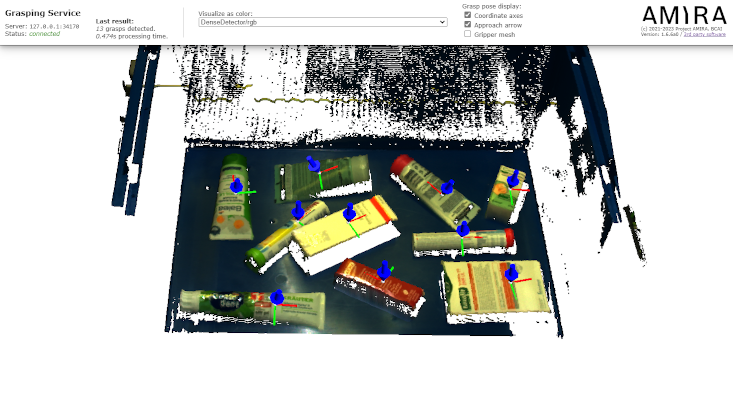}
\includegraphics[width=.19\linewidth,trim=75 0 100 50,clip]{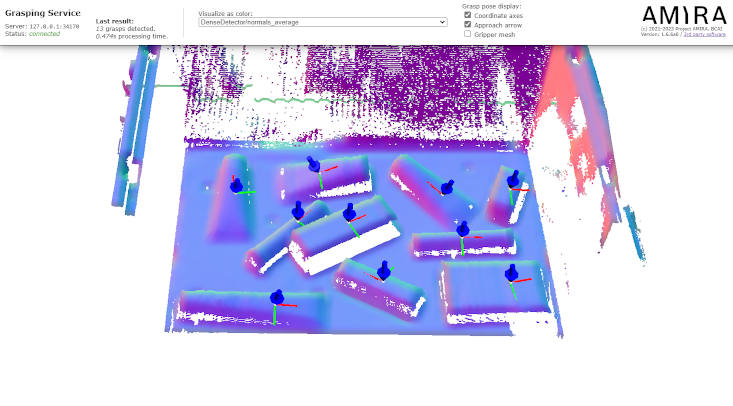}
\includegraphics[width=.19\linewidth,trim=75 0 100 50,clip]{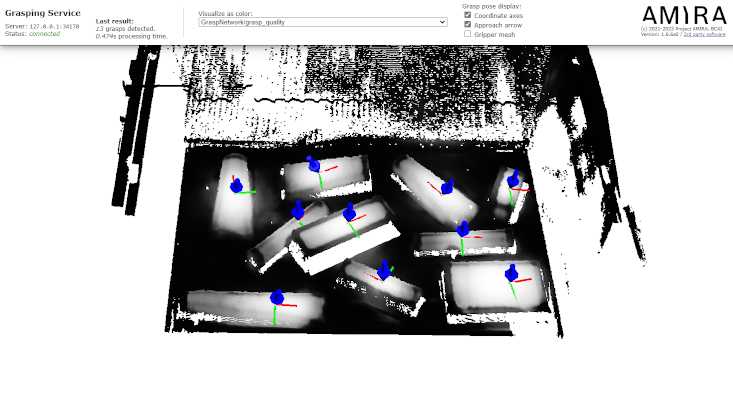}
\includegraphics[width=.19\linewidth,trim=75 0 100 50,clip]{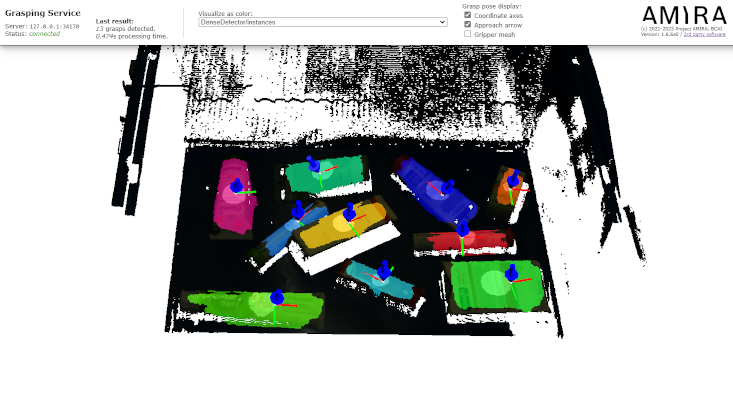}
\includegraphics[width=.19\linewidth,trim=75 0 100 50,clip]{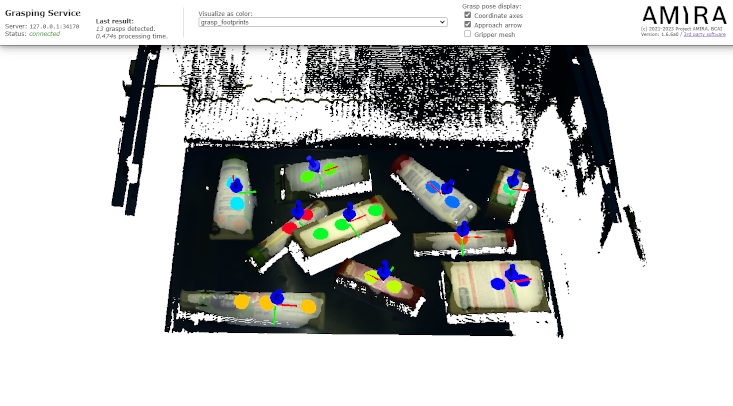}
\vspace*{-1em}  
\caption{Summary of intermediate steps of grasp detection on an example scene from an application (outside of the training distribution). From left to right: RGB input, calculated surface normals, inferred grasp quality, clustered graspable objects, and matched footprints.}
\vspace*{-1em}
\end{figure*}

However, accumulating grasp quality like this leaves one issue which we denote by the term "edge wrapping":
Consider a box which can be grasped well on two sides with different orientations, but not at the edge between these sides.
Using a gripper footprint with two suction cups that have enough space between the cups might now match one suction cup to one side and the other cup to the other side.
This would result in a grasp with a high theoretical graspability but which will fail in practice.
A similar example for edge wrapping is shown in Fig.~\ref{fig:footprint_edge_wrapping}.

Edge wrapping again motivates the use of normal vectors for avoiding infeasible grasp proposals.
We therefore perform another convolution with the same kernel $F$ over the three-channel image of normal vectors $S_\mathrm{Normals}$ to compute the standard deviation of normal vectors for each of the respective gripper footprint areas as
\begin{align}
F_\mathrm{Std} := \left| F \ast S_\mathrm{Normals}^2 - \left(F \ast S_\mathrm{Normals}\right)^2 \right|^\frac{1}{2}.
\end{align}
The product of the accumulated grasp quality $F \ast Q$ with the inverse of the above standard deviation $F_\mathrm{Std}$ then denotes the grasp feasibility.

Finally, we compute three single-channel pixel-wise results of this operation. 
The first result $O_\mathrm{Type}$ gives the pixel-wise gripper type and the second result $O_\mathrm{Rot}$ denotes the respective gripper rotation,
both given by the pixel-wise argmax over all $n_\mathrm{r} \cdot n_\mathrm{f}$ channels of the convolution result.
The last result is given by a pixel-wise grasp quality $O_\mathrm{Q}$, calculated as the $\max$ over all channels of the convolution result.
This is similar to the previously computed grasp quality $Q$, but $O_\mathrm{Q}$ now denotes for each pixel how feasible the best possible grasp configuration would be at that pixel in contrast to how well a pixel is suitable for being grasped.

\subsection{Pixel-to-Pose Transformation}\label{sec:transformation}

To obtain a list of grasps from the previous pixel-wise results, grasp quality values $Q$ are clustered such that clusters correspond to graspable areas of objects and pixels near the cluster center with highest $O_\mathrm{Q}$ are selected for grasping.
This has shown to improve robustness compared to directly using the pixels with highest values and ensures that grasps are detected for all objects that receive high (but not necessarily the highest) grasp quality values in a scene.

For each selected pixel $(u,v)$, the grasp pose position is then given by the respective point in the ordered point cloud $x, y, z := S_\mathrm{Pts}(u, v)$.
The roll and pitch rotations (around the $x$- and $y$-axes of the gripper) are determined by the surface of the object and are thus given by the negative normal vector at the respective pixel $\alpha, \beta := -S_\mathrm{Normals}(u, v)$.
The yaw rotation around the gripper axis is given by the gripper rotation determined during footprint matching $\gamma := O_\mathrm{Rot}(u, v)$.
Finally, the gripper type of the grasp is given by the determined footprint $t = O_\mathrm{Type}(u, v)$.

\section{Experiments}
\label{sec:experiments}
We perform two different types of experiments to evaluate the performance of our approach.
First, we evaluate grasp quality prediction performance by detecting single-suction grasps for an evaluation dataset and compare it with related methods.
Second, we demonstrate our multi-suction grasp detection on an industrial bin picking robot cell.

\subsection{Single-Suction Grasp Comparison}
Due to the lack of a comparable multi-suction gripper work, we determine single-suction grasps on an annotated reference dataset from the target bin picking cell and quantify the performance from the given pixel-wise ground-truth grasp success.
We compare our work to the following two methods for single-suction grasp detection:

\textbf{Dex-Net.} \citet{satish2019policy} propose a fully convolutional grasp quality CNN (FC-GQ-CNN) for grasp prediction.
We use the pretrained model FCGQCNN-4.0-SUCTION provided by the authors, which was trained on synthetic depth images of objects in clutter with parameters for a Photoneo PhoXi S camera.
Published values are based on a very specific combination and placement of RGB-D camera and gripper.
Therefore, we apply static cropping close to the bounding box of the bin and a depth value shift to make the depth images similar to the provided example images of the authors.
Input images are resized to $640 \times 480$.

\textbf{Zeng et al.} \cite{zeng2022robotic} introduce a multi-modal grasping framework.
They use a separate FCN with residual connections that generates a suctionability map for each pixel.
The framework combines RGB and depth data of size $640\times 480$ using two pre-trained ResNet-101 towers, then concatenates the output features to predict suction affordances.
In this paper, we evaluate the framework using two variations:
First, we replicated and trained the network with the original dataset of the authors.
Second, we retrained on a combination of the authors' dataset and our own dataset with labels and loss as in Sec.~\ref{sec:training}.

\begin{figure}[b]
	\vspace*{-1em}
	\centering
	\includegraphics[width=.42\linewidth]{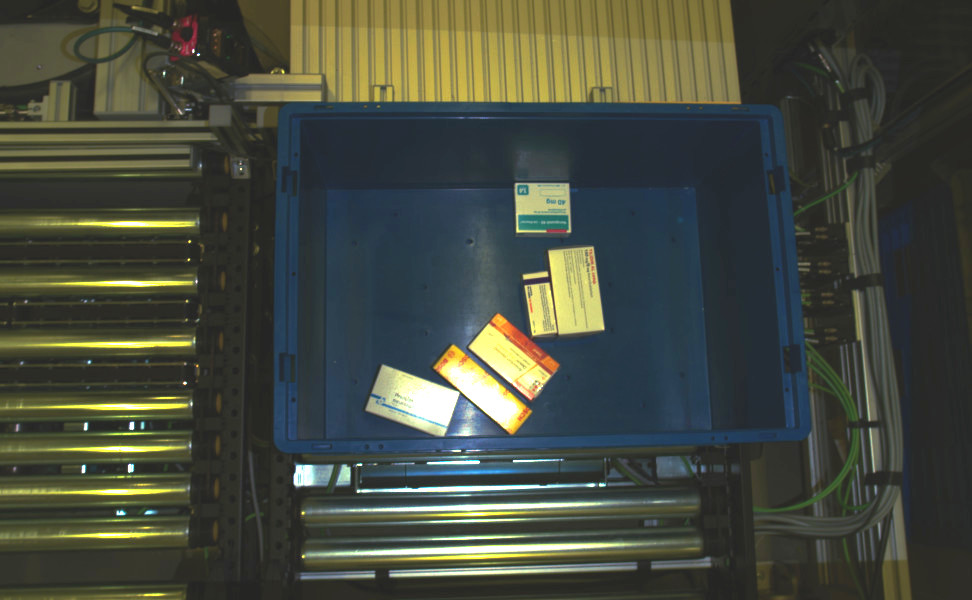}
	\includegraphics[width=.42\linewidth]{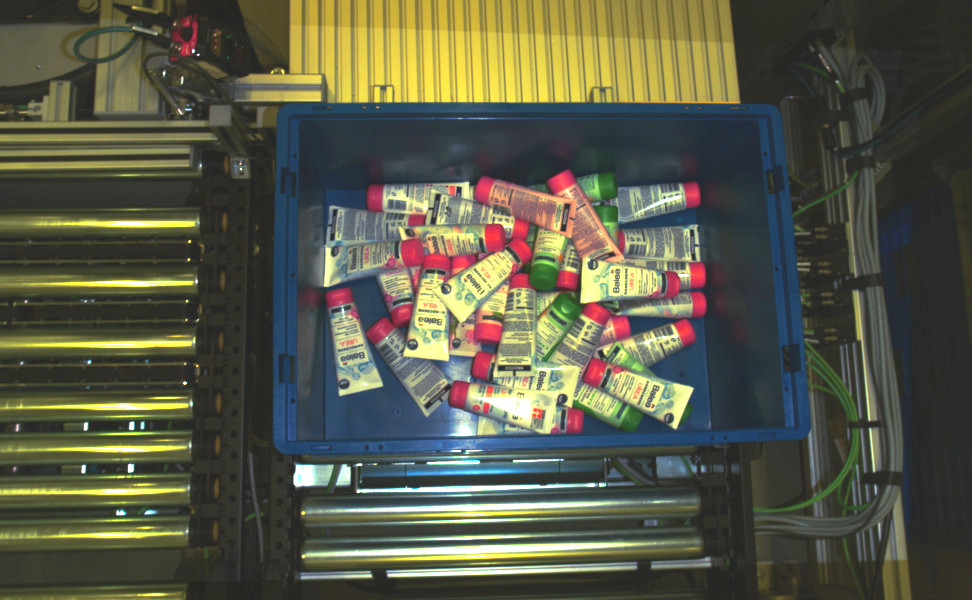}\\[.3em]
	\includegraphics[width=.42\linewidth]{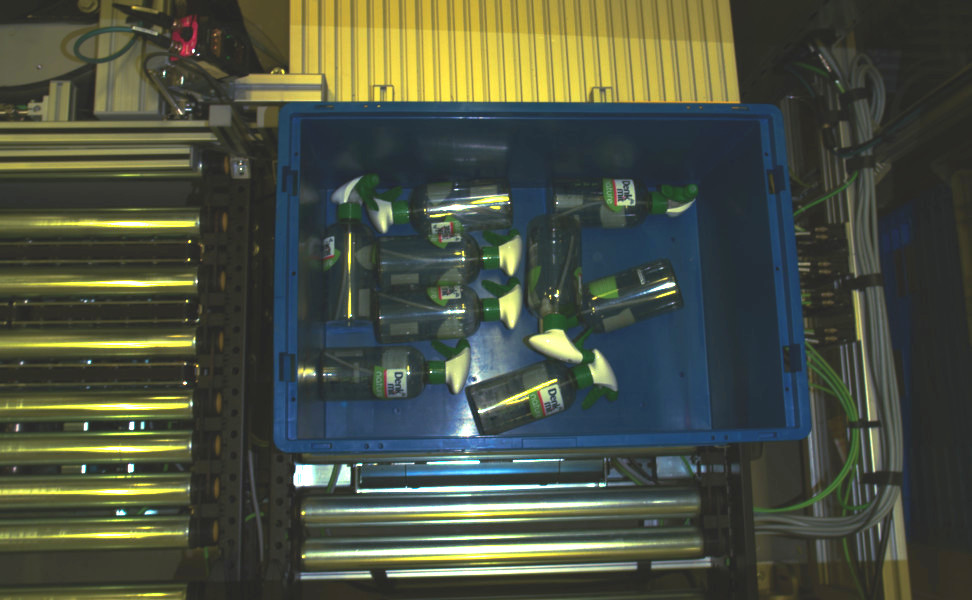}
	\includegraphics[width=.42\linewidth]{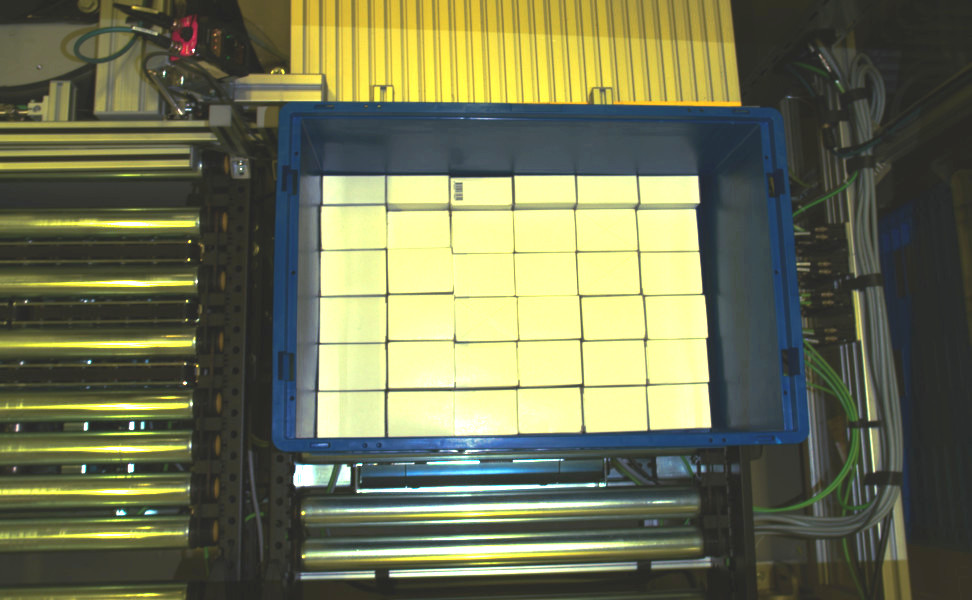}\\
	\caption{Examples for categories Simple (\textit{top left}), Typical (\textit{top right}) and Complex (\textit{bottom row}) of the evaluation dataset.}
	\label{fig:datasets}
\end{figure}

The dataset for evaluation is created from eleven types of common objects, see Fig.~\ref{fig:datasets} for examples.
For detailed results, we split the dataset into three levels of difficulty.
\textbf{Simple:} Bins filled with a small number of objects such as boxes or cylinders.
\textbf{Typical:} Bins containing a heap of a larger number of the objects.
\textbf{Complex:} Bins filled with a large number of challenging objects like partially transparent objects, stacked and texture-less boxes, or blister packs.

Each category consists of ten RGB-D images and corresponding ground-truth pixel-wise grasp sucess.
The grasp success is based on manually annotated object instance masks, combined with a similar method as described in Sec.~\ref{sec:training} for consideration of surface and weight, but with higher emphasis on the distance to the center.
To achieve meaningful results, we manually reviewed and corrected the grasp quality to ensure reasonable ground-truth labels.

For each scene, each method is queried for the top 20 grasps based on their grasp quality prediction.
Results are listed in Tab.~\ref{tab:dataonly_results}.
We state for each category 
as \emph{Quality} the average ground-truth grasp quality of feasible grasps (between $0$ and $1$),
as \emph{Success} the percentage of feasible grasps among predicted grasps (non-zero grasp quality),
as \emph{Objects} the percentage of objects for which grasps are detected (counting at most 20 objects per scene due to top 20 grasps),
as \emph{Multi} the average number of grasps predicted for the same object (closer to one is better),
and as \emph{None} the percentage of scenes without a single feasible grasp.

\begin{table}[t]
	\centering
	\caption{Performance of the approaches on our evaluation dataset. Our method achieves a competitive performance also for single-suction grasps and distributes them over objects.}
	\begin{tabular}{lrrrrr}
	\toprule
	\textbf{Simple} &        Quality &         Success &         Objects &         Multi &         None \\ \midrule
	Dex-Net         &          0.810 &          78.0\% &          59.3\% &          4.86 & \textbf{0\%} \\
	Zeng et al.     &          0.602 &          68.5\% &          21.9\% &         14.33 &         10\% \\
	--- finetuned   &          0.610 & \textbf{86.0\%} &          22.8\% &         15.78 &         10\% \\
	Ours            & \textbf{0.894} &          85.8\% & \textbf{82.2\%} & \textbf{1.07} & \textbf{0\%} \\ \bottomrule
\end{tabular}
\\[.2em]
\begin{tabular}{lrrrrr}
	\toprule
	\textbf{Typical} &        Quality &         Success &         Objects &         Multi &         None \\ \midrule
	Dex-Net          &          0.821 &          84.5\% &          32.0\% &          3.18 & \textbf{0\%} \\
	Zeng et al.      &          0.571 &          51.5\% &           6.5\% &         12.72 &         30\% \\
	--- finetuned    &          0.547 &          70.5\% &           6.0\% &         14.80 &         20\% \\
	Ours             & \textbf{0.909} & \textbf{90.1\%} & \textbf{84.5\%} & \textbf{1.01} & \textbf{0\%} \\ \bottomrule
\end{tabular}
\\[.2em]
\begin{tabular}{lrrrrr}
	\toprule
	\textbf{Complex} &        Quality &         Success &         Objects &         Multi &         None \\ \midrule
	Dex-Net          &          0.584 &          36.5\% &          15.8\% &          4.55 &         10\% \\
	Zeng et al.      &          0.299 &          23.5\% &           3.9\% &         14.17 &         50\% \\
	--- finetuned    &          0.601 & \textbf{81.5\%} &          10.4\% &         11.54 &         10\% \\
	Ours             & \textbf{0.790} &          56.4\% & \textbf{28.2\%} & \textbf{1.36} & \textbf{0\%} \\ \bottomrule
\end{tabular}
\\[.2em]

	\label{tab:dataonly_results}
	\vspace*{-1em}
\end{table}%

It can be seen that our method performs well for the considered application, which motivates its use for multi-suction grasp detection.
In particular, it can be observed that our method spreads grasps across objects compared to Dex-Net and Zeng et al., which often predict multiple alternatives for the same object.
Especially Zeng et al. achieve a presumably high success rate, but do so by detecting multiple close grasps per object which has limited usefulness in practice.
One contribution to this difference likely is the clustering of surface graspability for deriving poses (instead of directly predicting grasp success for pixels), as important for robustly placing multi-suction grippers.

We also achieve a high average grasp quality and while this does not directly indicate the feasibility for multi-suction grasps, it suggests that considering a larger footprint geometry also improves the overall quality of single-suction grasps.
This might be because the larger gripper geometry forces resulting grasp poses to be more consistently located on highly graspable areas and close towards object centers.

Finally, we would like to emphasize that we do not claim to generally outperform any of the other methods.
We verified that our method works sufficiently well on the intended use case and object portfolio compared to related work.
In general, applying the proposed gripper footprint optimization can also be done on grasp quality maps predicted by other methods.
However, we observed that it works most robustly for the procedure presented in this paper.
Still, especially Dex-Net shows a remarkable performance in the presented evaluation, considering that we were able to directly apply pretrained weights with minimal finetuning or preprocessing.

\subsection{Real-World Multi-Suction Grasp Experiments}

We finally demonstrate our approach by detecting poses for multi-suction grasps in the target application of industrial bin picking.
The experiment is performed on an industry-grade robotic bin picking cell as shown in Fig.~\ref{fig:bin_picking_setup}.
The cell is equipped with two \emph{Zivid Two} overhead RGB-D cameras with a resolution of $1944 \times 1200$ pixels, located around \SI{1.2}{\meter} above the bins.
It includes a \emph{Kuka KR10 R1420} robot with a custom-made linear gripper with multiple suction cups and activation schemes,
the same as shown in Fig.~\ref{fig:footprints} (left).

The cell is operated by an industrial software stack based on the \emph{Nexeed Automation}
framework with grasp poses provided by a \emph{PyTorch}-based implementation of our method that runs on a dedicated IPC of the cell for perception.
The IPC runs Ubuntu \num{20.04} and is equipped with an \emph{Intel Xeon W-1290T} CPU and a \emph{Nvidia Quadro RTX 4000} GPU.

\begin{figure}[t]
	\centering
	\begin{tikzpicture}
	\node at (0, 0) {\includegraphics[width=.7\linewidth,trim=0 2 10 2,clip]{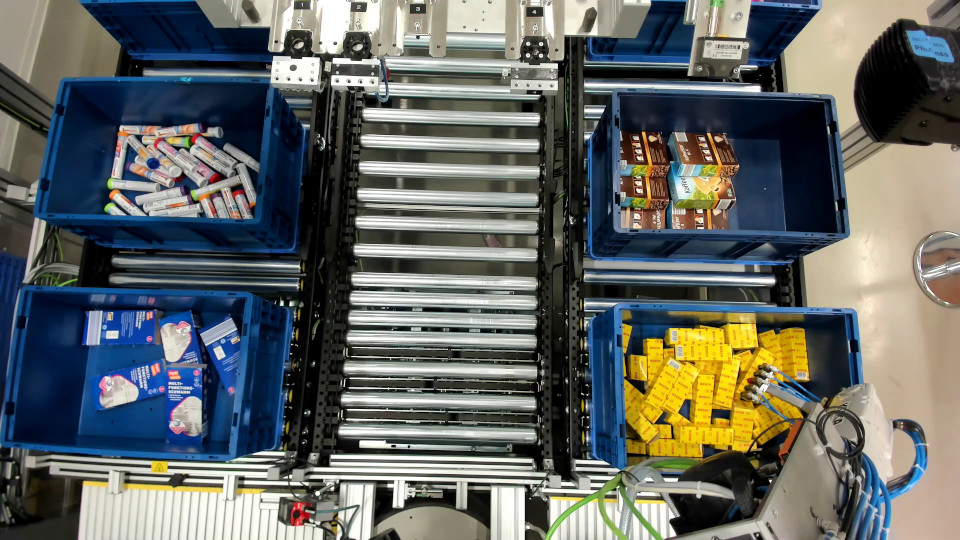}};
	\node at (-3.5, 0.55) {\includegraphics[width=.37\linewidth]{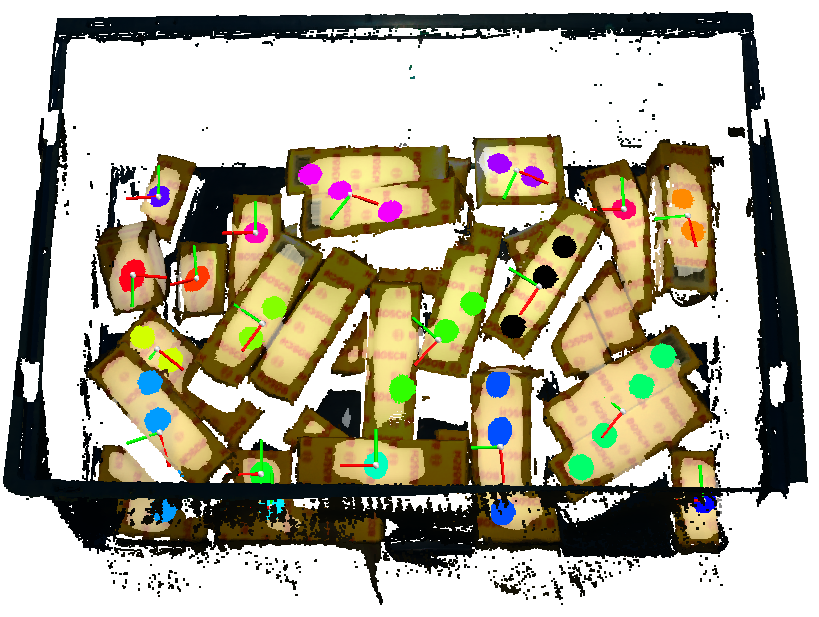}};
	\draw[->, color=white, line width=0.15em, shorten >=.4em,shorten <=.4em] (-2.05, 0.5) to (1, -0.7);
	\end{tikzpicture}
	\caption{Visualization of the grasps and corresponding footprints predicted by our approach for an exemplary scene. The executed grasp with selected three-cup footprint is colored in black in the visualization.}
	\label{fig:inno_experiments}
	\vspace*{-1em}
\end{figure}

A short video of our experiment is provided online\footnote{\vspace*{-1em}See experiment video: \url{https://youtu.be/UZikmSjQy3M}}.
Various object types similar to those from the grasp quality evaluation are provided in six different bins by the conveyor belt system.
We set an arbitrary sequence of picking orders in which the system composes deliveries from the available objects, simulating typical operations in a logistics center.

In the experiment, detecting multi-suction grasps requires on average \SI{628}{\milli\second}, of which grasp quality inference takes \SI{53}{\milli\second} and gripper footprint optimization takes \SI{372}{\milli\second} for the configured four footprints, a maximum rotation of \SI{180}{\degree} (due to symmetry) and a rotation resolution of \SI{5}{\degree}.
As in the comparison, each detection results in a list of up to 20 grasps from which subsequently, the path planner can select one to execute and considers the respective gripper activation.

Fig.~\ref{fig:inno_experiments} shows the grasp predictions and the performed grasp for one of the scenes from the video.
The selected grasp (black) places the footprint centrally on the object and fits three suction cups to increase the robustness.
The performed grasp matches the predicted footprint within the error margins of the overall system.

For the complete experiment, the system executed 38 grasps with three failures, i.e., a success rate of $92\%$.
In case of a failed grasp attempt, the system automatically executes the next feasible grasp pose.
Note that there are no pixel-wise ground-truth labels available in such real-world runs, thus we cannot determine all metrics as in Tab.~\ref{tab:dataonly_results} for the evaluation dataset.
Still, this qualitative experiment verifies practical applicability of the method in an industrial setting, and metrics such as the success rate match the expectations from our previous evaluation.

Overall, it can be seen from the video that choosing multi-suction grasps for larger objects indeed leads to an improved robustness for the grasps.
Still, we also observe that the overall system often selects grasps with a single suction cup.
This can be attributed to the fact that the path planner is allowed to freely rotate poses for the single-suction grasps due to being rotation-symmetric, which significantly increases the likelihood to find a feasible trajectory.

For smaller objects, the identified graspable areas are sometimes too small for fitting a footprint, a failure case that can be observed for the narrow cylindrical objects where once in the video, only a single grasp pose is detected but deemed infeasible by the path planner.
Finally, one concern was that the simple projection of a 2D footprint onto the scene surface might create projection issues on strongly tilted or bent surfaces, but we did not observe practical issues resulting from it.

\section{Conclusions}

In this paper, we proposed a method for detecting and optimizing multi-suction grasp poses for bin picking tasks
based on a model-free, gripper-agnostic prediction of pixel-wise graspability values.
In addition, we presented an automated procedure for labeling of images for supervised training of the grasp quality network, allowing for a trade-off between annotation quality and labeling effort.
For optimizing the selection of an activation pattern and the orientation of a multi-suction grasp, we described a procedure based on a convolution of grasp quality and surface normals with gripper-specific footprints.

In our evaluation and real-world experiments, we observed that the approach reliably predicts poses for one or more suction cups in a realistic setting, leading to feasible and robust grasps performed by the system.
To address a broader portfolio of objects and surface properties, future work can include multi-channel grasp quality predictions to denote different surface requirements for gripper selection.
On the system side, future work may allow for a closer integration of grasp optimization and motion planning.

\bibliographystyle{plainnat}
\small
\bibliography{bibliography}

\end{document}